\documentclass{article}

\PassOptionsToPackage{numbers, compress}{natbib}
\usepackage[final,main]{neurips_2026}

\usepackage[utf8]{inputenc}
\usepackage[T1]{fontenc}
\usepackage{hyperref}
\usepackage{url}
\usepackage{booktabs}
\usepackage{amsfonts}
\usepackage{amsmath}
\usepackage{nicefrac}
\usepackage{microtype}
\usepackage{xcolor}
\usepackage{graphicx}
\usepackage{multirow}
\usepackage{makecell}
\usepackage{array}
\usepackage{caption}
\usepackage{subcaption}
\usepackage{pifont}
\usepackage{xspace}
\usepackage{algorithm}
\usepackage{algpseudocode}

\newcommand{\method}{PPMem\xspace}

\title{Persistent Planar Memory for Video World Models}

\author{%
  \textbf{Yuze He}$^{1,2}$ \quad \textbf{Bin Tan}$^{2}$ \quad \textbf{Zelin Gao}$^{3,2}$ \quad \textbf{Yujun Shen}$^{2}$ \\
  \textbf{Yong-Jin Liu}$^{1,}$\thanks{Corresponding authors.} \quad \textbf{Nan Xue}$^{2,*}$ \quad \textbf{Yinghao Xu}$^{3,2}$ \\[0.4em]
  \normalfont $^{1}$Tsinghua University \quad $^{2}$Ant Group \quad $^{3}$HKUST
}

\begin{document}
\maketitle

\begin{abstract}
Video world models that generate and explore environments at scale need a persistent 3D memory of what they have produced. Plane primitives offer a natural candidate: a compact set of oriented planes captures the dominant structural elements ({\em e.g.,} walls, floors, facades) of indoor and outdoor scenes, while requiring only 7 parameters per primitive.
We present \textbf{\method}, a video world model that adopts plane primitives as its persistent 3D memory. \method augments a video diffusion transformer with a planar prediction head and a Depth Token Fusion (DTF) module: each autoregressive chunk jointly produces RGB video and planar geometry, which merges into a persistent plane memory that conditions subsequent chunks via rendered depth. The resulting memory is three orders of magnitude more compact than per-pixel point clouds (${\sim}$100K primitives, ${\sim}$5\,MB for a full scene), geometrically explicit (metric depth, normals, exportable surfaces), cross-view consistent (jointly predicted by the shared backbone), and adaptive (refined by the merging procedure as generation proceeds).
On DL3DV and RealEstate10K, \method outperforms prior 3D-grounded world models in video quality while jointly producing metric-scale depth, surface normals, and exportable planar geometry, sustaining coherence across hundreds of frames under large viewpoint changes. {\it Project page:} \url{https://hyzcluster.github.io/ppmem}.
\end{abstract}

\section{Introduction}
\label{sec:intro}

\begin{figure}[t]
  \centering
  \includegraphics[width=\linewidth]{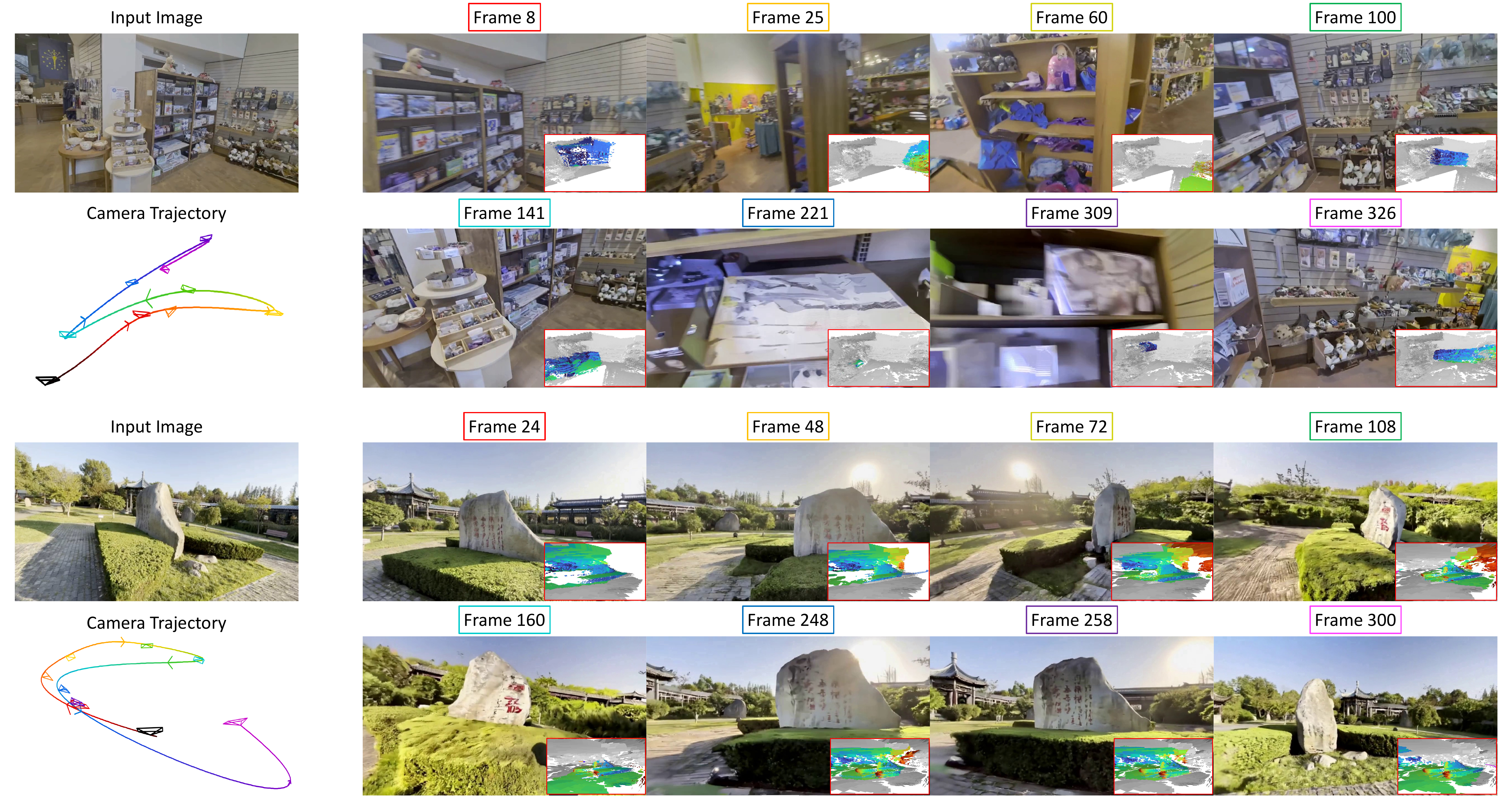}
  \caption{
    \textbf{Given a single input image and a camera trajectory, \method autoregressively generates video while accumulating a persistent plane-based 3D memory.}
    We show two DL3DV scenes (indoor, outdoor) with generated frames at selected indices. Insets visualize the plane pool at each frame, colored by the current view frustum. The plane geometry grows to cover the full explored scene, serving as persistent spatial memory for subsequent generation.
  }
  \label{fig:teaser}
\end{figure}

Humans build the mental representation of their world through structural abstraction, decomposing scenes into points, lines, and planes.
Among the three, planes carry disproportionate weight in human-made environments: walls, floors, ceilings, tabletops, building facades, and ground surfaces~\cite{coughlan1999manhattan,debevec1996modeling} can all be approximated by only a small set of planes.
A model that generates entire worlds, not just short clips, needs an analogous compact, structured memory of what it has produced.
Plane primitives are a natural candidate for this role.
Each primitive is parameterized by just 7 numbers (a center depth, two rectangular extents, and a rotation quaternion), which explicitly represents 3D geometric structure, and is orders of magnitude more compact than per-pixel representations.
This raises a natural question: \emph{can plane primitives serve as a video world model's representation, acting as a compact yet expressive persistent memory?}

Video world modeling has emerged as a central frontier of AI, with systems that generate, simulate, and explore environments at scale.
Existing approaches roughly fall into two camps.
\textbf{Pure-video world models}, such as Genie~3~\cite{genieteam2025genie3} and LingBot-world~\cite{gao2026lingbot}, train end-to-end on massive video corpora and rely on implicit attention and scale to maintain coherence. They achieve remarkable visual fidelity but expose no explicit 3D state for downstream conditioning, editing, or export.
\textbf{3D-grounded video world models}, the direct comparison group for \method, augment diffusion with explicit 3D geometric memory: a 2-frame point-cloud buffer (Gen3C~\cite{ren2025gen3c}), an oriented-surfel store (VMem~\cite{li2025vmem}), a TSDF volume paired with a keyframe pool (SPMem~\cite{wu2025video}), or a raw-frame retrieval bank (Context as Memory~\cite{yu2025context}, WorldMem~\cite{xiao2025worldmem}).
A related line, Voyager~\cite{huang2025voyager}, jointly generates RGB and depth video and re-projects the generated depth into a point-cloud cache for conditioning subsequent chunks.
Yet no existing 3D-grounded representation simultaneously exhibits the four properties that an explicit, persistent 3D state demands: \textbf{compactness} (orders of magnitude fewer primitives than dense representations), \textbf{geometric explicitness} (metric, rasterizable, exportable), \textbf{cross-view consistency} (jointly predicted by the generation backbone rather than assembled per frame), and \textbf{adaptivity} (refining the representation as generation proceeds).

We present \textbf{\method}, a video world model that uses plane primitives as its underlying 3D structure, accumulating them in a \textbf{persistent plane memory} for long-context video generation.
Concretely, \method augments a video diffusion transformer (DiT) with two additions: a planar prediction head adapted from Plana3R~\cite{liu2025plana3r} that regresses plane parameters per chunk, and a \textbf{Depth Token Fusion} (DTF) module that re-injects depth rendered from the memory into the DiT features.
The memory plays a dual role: as the model's structured 3D output (metric-scale depth, surface normals, and exportable planar geometry), and as persistent spatial conditioning for subsequent generation.
Plane primitives are particularly well-suited for this role: after merging, ${\sim}$100K planes cover a full scene three orders of magnitude more compactly than per-pixel point clouds (Table~\ref{tab:rep_compare}), and are fully differentiable through rectangular Gaussian rasterization~\cite{liu2025plana3r}.
Under large viewpoint changes, \method sustains coherent, metric-grounded video generation across hundreds of frames (Figure~\ref{fig:teaser}).

Compared with Gen3C, Voyager, VMem, and SPMem, \method achieves state-of-the-art video quality on DL3DV and RealEstate10K while accumulating under 10\,MB of 3D memory.

\textbf{Our contributions are:}
\begin{itemize}
\item We propose plane primitives as a 3D representation for video world models, accumulated in a persistent plane memory via cross-chunk geometric merging --- to our knowledge, the first such memory that is simultaneously compact, geometrically explicit, cross-view consistent, and adaptive.
\item We instantiate \method on a 5B Wan-2.2 DiT with a Depth Token Fusion (DTF) module that re-injects rendered plane-pool depth into the generation process, enabling joint generation of RGB, metric depth, surface normals, and planar geometry in a single framework.
\item We benchmark \method against Gen3C, Voyager, VMem, and SPMem on short-context and long-context evaluations across various datasets, demonstrating sustained visual and geometric coherence across hundreds of frames under large viewpoint changes.
\end{itemize}

\section{Related Work}
\label{sec:related}

\paragraph{Camera-controlled video generation.}
Recent video diffusion models with camera control~\cite{ren2025gen3c,bai2025recammaster,yu2025trajectorycrafter,he2024cameractrl,he2025cameractrl,wang2024motionctrl,huang2025voyager} generate impressive short clips of 14--121 frames. However, extending generation to longer sequences reveals a \emph{representation gap}: most methods either lack a persistent 3D state entirely (relying on camera embeddings or frame-level conditioning) or maintain only minimal geometric state. We organize prior work below by the \emph{level of 3D representation} they maintain, highlighting a progressive trend toward more structured representations.

\paragraph{From depth maps to point clouds.}
Depth estimation provides geometric cues for video world models~\cite{xu2026depthfoundation}.
At the simplest level, Gen3C~\cite{ren2025gen3c} maintains a 2-frame point buffer (the seed frame and the most recent chunk-boundary frame, whose monocular depth is re-estimated and aligned to the cache at each autoregressive step); these 3D points are forward-warped as conditioning for the next generation chunk. This provides per-chunk geometric grounding but carries no persistent structure beyond two frames: all intermediate geometry is overwritten. Sora3R~\cite{mai2025can} predicts dense 4D pointmaps but is restricted to short sequences. At this level, depth and point representations are per-pixel and unstructured, offering no abstraction beyond raw geometry.

\paragraph{Surfels, TSDF, and frame-level memory.}
A step up in structure, VMem~\cite{li2025vmem} reconstructs oriented surfels via CUT3R~\cite{wang2025continuous} during autoregressive generation and uses them to index relevant past views; however, the surfel count grows with each generated frame. SPMem~\cite{wu2025video} builds a TSDF volume as spatial memory alongside an episodic keyframe pool; extending to long sequences requires re-estimating depth and re-integrating all accumulated frames into the volume at each step, with cost scaling linearly in sequence length. Context as Memory~\cite{yu2025context} retrieves raw past frames by FOV overlap, and WorldMem~\cite{xiao2025worldmem} uses cross-attention over a frame bank with relative Pl\"ucker embeddings; both operate at the frame level without maintaining explicit 3D geometry.

\paragraph{Joint depth and 3D generation.}
A complementary line of work produces geometry alongside or from video. Voyager~\cite{huang2025voyager} jointly generates video and depth via shared diffusion and proposes re-projecting generated depth into a point-cloud cache for extended generation; its conditioning depth is normalized per-frame to relative disparity. IDCNet~\cite{liu2025idcnet} outputs metric depth with camera control. GeometryCrafter~\cite{xu2025geometrycrafter} is a post-hoc tool that estimates temporally coherent point maps from \emph{existing} video without generating new content. ViewCrafter~\cite{yu2024viewcrafter} renders an externally estimated point cloud (DUSt3R~\cite{wang2024dust3r}) at target viewpoints and uses diffusion to inpaint holes, consuming 3D rather than predicting it. FlashWorld~\cite{li2025flashworld} extends Wan~2.2~\cite{wan2025wan} with a 3D Gaussian decoder for joint multi-view generation, but repurposes the temporal dimension as a multi-view axis to produce \emph{static} 3D scenes rather than temporal video sequences. None of these methods accumulate a compact, structured 3D representation that serves as persistent memory for long-context video generation.

\paragraph{Plane-based 3D reconstruction and our position.}
Learning-based plane detection progresses from single-view segmentation (PlaneNet~\cite{liu2018planenet}, PlaneRCNN~\cite{liu2019planercnn}, PlaneTR~\cite{tan2021planetr}) to differentiable multi-view reconstruction: AlphaTablets~\cite{he2024alphatablets} optimizes alpha-composited planar primitives from monocular video, and PlanarSplatting~\cite{tan2025planarsplatting} extends 3D Gaussian splatting~\cite{wu2024recent} with planar constraints. Plana3R~\cite{liu2025plana3r} performs \emph{feed-forward} planar splatting via differentiable rectangular Gaussian rasterization, predicting metric-scale plane primitives from unposed image pairs in a single pass without per-scene optimization. All of these methods are \emph{reconstruction-only}: they extract geometry from already-observed images. \method repurposes planar prediction as \emph{persistent spatial memory within a generative model}: planes are jointly predicted alongside each generated video chunk by a shared diffusion backbone, and the accumulated plane pool feeds back as geometric conditioning for subsequent chunks. This bidirectional loop (generation $\to$ planes $\to$ conditioning $\to$ generation) has no counterpart in prior plane reconstruction work. We compare against alternative 3D representations in \S\ref{sec:method_repr}.

\section{\method: Persistent Planar Memory}
\label{sec:method}

\method uses plane primitives as a compact, persistent 3D memory for autoregressive video generation. Unlike point clouds, surfels, or TSDF volumes used by prior 3D-grounded world models, plane primitives are predicted \emph{internally} by the generation backbone (requiring no external depth estimator), yield cross-view consistent geometry by construction, and compress an entire scene into ${\sim}$100K primitives --- three orders of magnitude fewer than dense point clouds (Table~\ref{tab:rep_compare}). Each generated chunk jointly predicts video and a set of oriented rectangles in 3D, which are accumulated into a global pool via geometric merging; the pool is then rendered and fed back as depth conditioning for subsequent chunks. We describe the representation and memory mechanism (\S\ref{sec:method_repr}), the model architecture (\S\ref{sec:method_arch}), and training (\S\ref{sec:method_train}) below.

\subsection{Plane Primitives and Persistent Memory}
\label{sec:method_repr}

A plane primitive in our framework is a parametric oriented rectangle in 3D space, defined by 7 parameters: a \emph{center depth} $z$ along the camera ray (from which the 3D center $\mathbf{x}$ is derived via the camera intrinsics), two \emph{rectangular extents} $(s_x, s_y)$, and a \emph{rotation quaternion} $q \in \mathbb{R}^4$ whose third rotation-matrix column gives the surface normal $\mathbf{n}$. Each primitive is rendered via differentiable rectangular Gaussian rasterization~\cite{liu2025plana3r}, producing depth and normal maps. At inference, predicted primitives are transformed to world coordinates and accumulated into the persistent plane pool, which after merging typically contains ${\sim}$100K primitives (${\sim}$5\,MB) representing the full explored scene. Beyond compactness and internal predictability (Table~\ref{tab:rep_compare}), plane parameters can be merged across autoregressive steps via simple geometric criteria (occlusion voting, coplanarity, distance), making them well-suited as a cross-chunk memory.

\begin{table}[t]
  \centering
  \caption{
    \textbf{Properties of 3D representations for cross-chunk video memory.}
    We compare along four aspects: whether geometry is predicted \emph{internally} by the generation model or externally, whether predictions are cross-view consistent, the number of persistent 3D primitives, and whether the representation captures structured surfaces. Primitive counts are estimates at ${\sim}$400 frames, $576{\times}1024$.
  }
  \label{tab:rep_compare}
  \small
  \resizebox{\linewidth}{!}{%
  \begin{tabular}{lcccc}
    \toprule
    & Dense PC~\cite{ren2025gen3c} & Surfels~\cite{li2025vmem} & TSDF$\to$PC~\cite{wu2025video} & \textbf{Planes (\method)} \\
    \midrule
    3D source         & External (MoGe~\cite{wang2025moge})  & External (CUT3R~\cite{wang2025continuous})  & External (DA3~\cite{lin2025depth}) & \textbf{Internal (joint head)} \\
    Cross-view consistent & \ding{55}    & \ding{55}         & partial        & \textbf{\ding{51}} \\
    Primitives (accumulated) & ${\sim}$210M$^\dagger$ & ${\sim}$250K  & ${\sim}$1--2M & \textbf{${\sim}$100K} \\
    Structured surfaces & \ding{55}      & \ding{55}         & \ding{55}      & \textbf{\ding{51}} \\
    \bottomrule
  \end{tabular}}

  \vspace{0.4em}
  {\footnotesize
  $^\dagger$Theoretical total if all per-pixel points were accumulated.
  }
\end{table}

\paragraph{Persistent plane pool.}
The persistent plane pool $\mathcal{P}=\{\mathbf{p}_1,\mathbf{p}_2,\ldots\}$ is a global, accumulating set of plane primitives across autoregressive chunks. After each chunk produces predicted planes $\mathcal{P}_{\text{new}}$, we merge them into $\mathcal{P}$ via the procedure below. Before the next chunk, we render $\mathcal{P}$ at the target camera poses and feed the rendered depth as conditioning to the generation model (detailed in \S\ref{sec:method_arch}). The full inference loop is summarized in Algorithm~\ref{alg:inference} (Appendix).

\paragraph{Plane pool maintenance.}
After each chunk, new planes are merged into $\mathcal{P}$ with two goals: remove erroneous pool planes exposed by newer observations, and add new planes only where they contribute genuine new geometry. Each plane carries an \textbf{observation quality score} $s = r / |\cos\theta|$, where $r$ is the Euclidean distance to the creating camera and $\theta$ is the angle between the viewing ray and the surface normal (lower is better: the camera is closer and views the surface more perpendicularly). Three mechanisms maintain pool quality:
\begin{enumerate}
  \itemsep0em
  \item \textbf{Occlusion deletion} (\emph{correcting the past}): pool planes that float in front of newly predicted surfaces are flagged via multi-camera voting and deleted.
  \item \textbf{Bidirectional quality dedup} (\emph{improving the present}): among nearby planes (within distance $\epsilon$), only the one with the better quality score survives --- both within a chunk (\emph{intra-step}, typically reducing count by ${\sim}80\%$) and across chunks (\emph{cross-step}).
  \item \textbf{Forward-looking prune} (\emph{anticipating the future}): pool planes that will be viewed from much closer in the next chunk ($r_{\text{next}} / r_{\text{created}} < \tau$) are removed proactively, as the upcoming chunk will produce higher-quality replacements. This uses only the next chunk's camera poses, which are supplied as camera-control inputs rather than predicted.
\end{enumerate}
Additional filtering stages are detailed in Algorithm~\ref{alg:merge} (Appendix). Surviving new planes are appended with metadata. Merging is local and does not require gradient-based optimization.

\paragraph{Pool-to-chunk conditioning.}
Before each new chunk, we render $\mathcal{P}$ at the target camera poses via differentiable rasterization, producing partial depth maps that condition the next chunk through two complementary paths: VAE-encoded as a latent channel-concatenated with the noisy video input, and at pixel resolution through token-level fusion (\S\ref{sec:method_arch}). Unlike the fixed-size buffers of prior work, $\mathcal{P}$ accumulates geometry across \emph{all} past chunks; its growth is controlled by merging, typically reaching ${\sim}$100K primitives (${\sim}$5\,MB) after ${\sim}$400 frames.

\subsection{Model Architecture}
\label{sec:method_arch}

\begin{figure}[t]
  \centering
  \includegraphics[width=\linewidth]{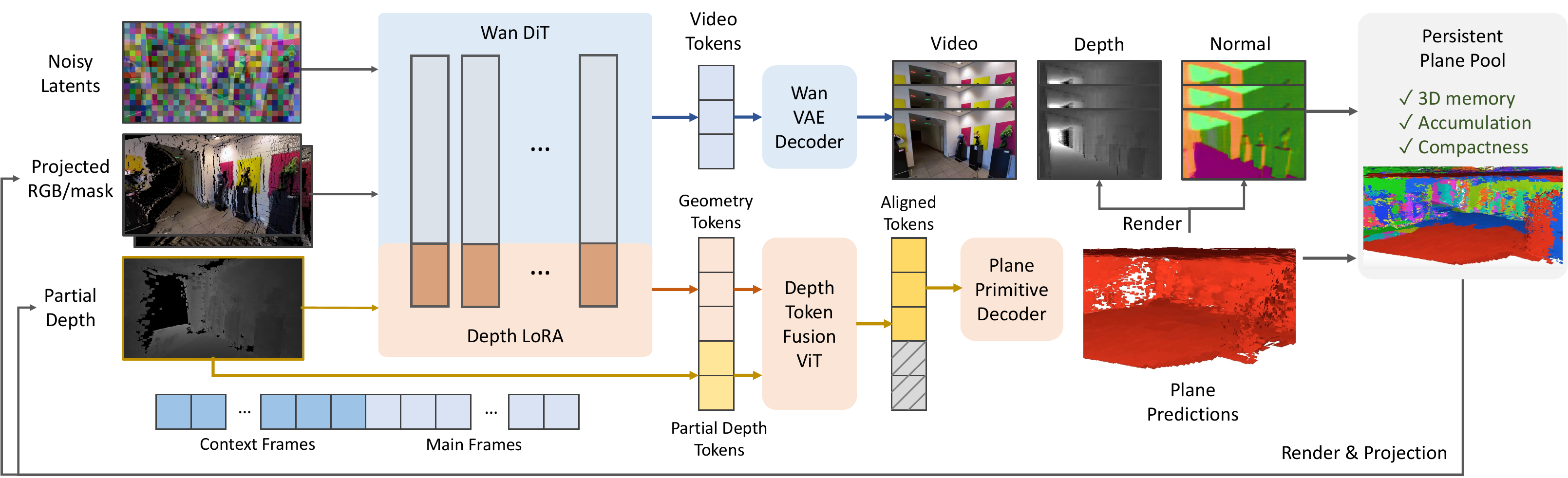}
  \caption{
    \textbf{\method architecture overview.} The Wan DiT receives noisy video latents channel-concatenated with projected source-view RGB/masks and partial depth rendered from the plane pool; context frames are prepended temporally. The DiT produces two token streams: \emph{video tokens} decoded by the Wan VAE into RGB outputs, and \emph{geometry tokens} adapted by geometry LoRA and fused with partial depth tokens via a ViT-based Depth Token Fusion module. The fused tokens are decoded into per-frame plane primitives, which merge into the persistent plane pool. The pool is rendered at the next chunk's target cameras and fed back as conditioning.
  }
  \label{fig:arch}
\end{figure}

\method builds on the Wan~2.2 video diffusion transformer~\cite{wan2025wan} with two architectural additions (Figure~\ref{fig:arch}): a \textbf{plane prediction head} that regresses plane primitives from DiT features, and a \textbf{Depth Token Fusion (DTF) module} that injects rendered plane-pool depth as token-level conditioning. At inference, \method generates video in ${\sim}$25-frame chunks: target camera poses are used to project source views and render the plane pool into geometric conditioning (projected RGB, visibility masks, and depth maps), which are channel-concatenated with the noisy latent and fed to the DiT. The camera is encoded entirely through projected views rather than explicit parameters, decoupling the model from specific intrinsics or trajectory scales.

\paragraph{Plane prediction head.}
\label{sec:method_head}
The head, adapted from Plana3R~\cite{liu2025plana3r}, regresses the 7 plane parameters (\S\ref{sec:method_repr}) from DiT features. Since the DiT operates on a coarsely-patchified latent grid ($H/16 \times W/16$), we first apply a \textbf{feature decoder} that upsamples by $2{\times}$ in the spatial dimension. The decoder is initialized from the Wan~2.2 VAE decoder's upsampling block, leveraging learned spatial priors~\cite{mai2025can}. Three small MLPs then regress center depth (in log-space over $[0.2, 277]$\,m), rectangular extents (in metric units), and rotation quaternion at each feature location. The predicted primitives are rendered via differentiable rectangular Gaussian rasterization~\cite{liu2025plana3r} and supervised against reconstructed metric depth (\S\ref{sec:method_train}).

\paragraph{Depth token fusion (DTF).}
\label{sec:method_dtf}
To inject the plane pool's geometric context into plane prediction, we use a DTF module (inspired by MDM~\cite{tan2026masked}) in the geometry pass (\S\ref{sec:method_train}). Partial depth maps rendered from $\mathcal{P}$ are tokenized via a $16{\times}16$ patch convolution and processed by a small ViT-Base encoder (${\sim}$87\,M parameters, ${\sim}$1.7\% of the 5B DiT), initialized from DINOv2-Base~\cite{oquab2023dinov2}; the encoded depth tokens are then concatenated with DiT feature tokens for fusion. Since rendered depth maps contain holes (unseen regions) and partially covered patches, we adopt a \textbf{three-tier masking} strategy before fusion: (i) \emph{fully invalid} patches (zero coverage) are always dropped; (ii) \emph{mixed} patches (partial coverage) are dropped with probability $\rho_{\text{mix}}{=}0.75$; (iii) \emph{fully valid} patches are randomly dropped to reach a target overall mask ratio in $[0.6, 0.9]$. This forces the DiT to remain robust to depth-conditioning sparsity while still benefiting from reliable conditioning signal.

\subsection{Training}
\label{sec:method_train}

We train \method in two stages.
\textbf{Stage~1} fine-tunes the Wan~2.2 backbone for camera-controlled video generation, conditioning on projected source views and context frames with the standard flow-matching diffusion loss~\cite{lipman2022flow}.
\textbf{Stage~2} freezes the video backbone and attaches the plane prediction head, DTF module, and feature decoder; training proceeds exclusively with a geometry loss:
\begin{equation}
  \mathcal{L} = \underbrace{\lambda_d \mathcal{L}_{\text{depth}} + \lambda_n \mathcal{L}_{\text{normal}}}_{\text{per-patch}} + \underbrace{\lambda_{d}^{r} \mathcal{L}_{\text{depth}}^{\text{rend}} + \lambda_{n}^{r} \mathcal{L}_{\text{normal}}^{\text{rend}}}_{\text{rasterization}} + \lambda_s \mathcal{L}_{\text{smooth}},
\end{equation}
where the per-patch terms supervise depth ($L_1$, log-space) and normals (cosine + $L_1$) per spatial token, the rasterization terms compare rendered planes against ground truth (enabled after 1000 warmup iterations), and $\mathcal{L}_{\text{smooth}}$ is a total-variation regularizer on rendered depth.
Default weights: $\lambda_d {=} 0.2$, $\lambda_n {=} 0.05$, $\lambda_d^r {=} 0.02$, $\lambda_n^r {=} 0.01$, $\lambda_s {=} 0.02$ (full configuration in Appendix~\ref{app:train_hp}).
We inject LoRA adapters~\cite{hu2022lora} on self-attention in all 30 DiT blocks, providing geometry-specific attention patterns atop the frozen video-generation weights.
At inference, video and geometry are fully decoupled: all denoising steps run with the frozen Stage-1 backbone (LoRA disabled), producing video at the original model's quality. A single additional forward pass with LoRA enabled extracts plane geometry from the final-step features.
Stage~1 runs for ${\sim}$7\,000 iterations and Stage~2 for ${\sim}$5\,000.

\section{Experiments}
\label{sec:exp}

\subsection{Setup}
\label{sec:exp_setup}

\textbf{Training data.} We train on a mixture of DL3DV~\cite{ling2024dl3dv} and RealEstate10K~\cite{zhou2018stereo} (80K+ clips) at $576{\times}1024$ resolution. For ablation experiments (Tables~\ref{tab:abl_repr}--\ref{tab:abl_decoder}), we train on a DL3DV indoor subset (${\sim}$5K clips) at $288{\times}512$. Both settings use metric depth from MoGe-2~\cite{wang2025moge2}, scale-aligned to VGGT~\cite{wang2025vggt} reconstructions (Appendix~\ref{app:data}).

\textbf{Test sets.} We evaluate on 100 DL3DV test scenes and 100 RealEstate10K (RE10K) clips (both disjoint from training); ablations use the DL3DV indoor subset (56 scenes). For long-context evaluation, we use the full DL3DV test sequences (300--400 frames each), which exhibit substantially larger camera motion than RE10K and stress the persistent memory mechanism.

\textbf{Baselines.} We compare against Gen3C~\cite{ren2025gen3c}, Voyager~\cite{huang2025voyager}, VMem~\cite{li2025vmem}, and SPMem~\cite{wu2025video}, all run with their public release code and default hyperparameters. All methods receive ground-truth camera trajectories and first-frame depth (or Pl\"ucker rays for VMem).

\textbf{Protocol.} Each method generates at its own native resolution and aspect ratio; input images and ground truth are center-cropped and resized to match, so that metrics reflect each method's best operating point. We report short-range results at $N{=}49$ and $N{=}121$ frames (Table~\ref{tab:short_range}).

\textbf{Metrics.} For short-range evaluation (Table~\ref{tab:short_range}), we report PSNR, LPIPS, CLIP similarity, and DINOv2 similarity. For long-context evaluation (Figure~\ref{fig:degradation}), we report CLIP and DINOv2 similarity as a function of frame index, which capture semantic and structural fidelity across large viewpoint changes. Per-frame depth accuracy against baselines is shown in Figure~\ref{fig:depth_and_dtf}(a); depth metrics --- absolute relative error (AbsRel $= |d - d^*|/d^*$), threshold accuracy ($\delta{<}1.25$), and mean normal angle error (Normal$^\circ$) --- are additionally reported in the ablation tables.

\textbf{Implementation.} \method is built on Wan~2.2 TI2V 5B~\cite{wan2025wan}. We use a per-frame VAE encoder (no temporal compression, 1:1 latent-to-frame mapping) to preserve per-frame geometric correspondence for plane prediction. Depth conditioning is log-normalized over $[0.2, 99]$\,m, while the plane head predicts log-depth over $[0.2, 277]$\,m. We train with learning rate $1{\times}10^{-5}$, classifier-free guidance dropout 0.1, and context-block dropout 0.1 (all context frames dropped together). Full hyperparameters are in Appendix Table~\ref{tab:hp_full}.

\subsection{Short-Range Comparison}
\label{sec:exp_short}

Table~\ref{tab:short_range} reports short-range video quality on DL3DV and RealEstate10K. \method achieves the best results across all metrics on both datasets and both frame budgets, despite using a smaller base model (Wan~2.2 5B) compared to Gen3C (Cosmos 7B) and Voyager (HunyuanVideo ${\sim}$13B). Table~\ref{tab:short_range} compares complete systems; Table~\ref{tab:abl_repr} isolates the memory mechanism by swapping it on the same Wan~2.2 generator.

Beyond video quality, \method jointly outputs metric-scale depth. Figure~\ref{fig:depth_and_dtf}(a) compares per-frame depth accuracy against Voyager --- the only other baseline with joint depth output --- over 121 frames on DL3DV. \method maintains lower AbsRel and higher $\delta{<}1.25$ accuracy than Voyager.

\begin{table}[t]
  \centering
  \caption{
    \textbf{Short-context comparison on DL3DV and RealEstate10K.}
    We report two frame budgets: $N{=}49$ (native chunk size for Voyager and SPMem; VMem uses 48) and $N{=}121$ (Gen3C native chunk). At $N{=}49$ most baselines are single-chunk; at $N{=}121$ Voyager, VMem, and SPMem all require autoregressive looping.
  }
  \label{tab:short_range}
  \small
  \setlength{\tabcolsep}{3pt}
  \resizebox{\linewidth}{!}{%
  \begin{tabular}{ll cccc cccc}
    \toprule
    & & \multicolumn{4}{c}{$N{=}49$} & \multicolumn{4}{c}{$N{=}121$} \\
    \cmidrule(lr){3-6}\cmidrule(lr){7-10}
    Dataset & Method & PSNR$\uparrow$ & LPIPS$\downarrow$ & CLIP$_\text{sim}\uparrow$ & DINOv2$_\text{sim}\uparrow$ & PSNR$\uparrow$ & LPIPS$\downarrow$ & CLIP$_\text{sim}\uparrow$ & DINOv2$_\text{sim}\uparrow$ \\
    \midrule
    \multirow{5}{*}{\rotatebox{90}{\small DL3DV}}
    & Gen3C~\cite{ren2025gen3c}     & \underline{15.74} & \underline{0.474} & 0.847 & \underline{0.710} & \underline{13.59} & \underline{0.565} & \underline{0.798} & \underline{0.585} \\
    & Voyager~\cite{huang2025voyager}   & 12.84 & 0.568 & 0.796 & 0.615 & 11.35 & 0.649 & 0.760 & 0.516 \\
    & VMem~\cite{li2025vmem}      & 11.88 & 0.636 & 0.781 & 0.477 & 10.87 & 0.722 & 0.696 & 0.274 \\
    & SPMem~\cite{wu2025video}     & 14.70 & 0.492 & \underline{0.853} & 0.703 & 13.03 & 0.607 & 0.788 & 0.536 \\
    & \textbf{\method (ours)} & \textbf{16.99} & \textbf{0.335} & \textbf{0.909} & \textbf{0.861} & \textbf{13.98} & \textbf{0.464} & \textbf{0.863} & \textbf{0.769} \\
    \midrule
    \multirow{5}{*}{\rotatebox{90}{\small RE10K}}
    & Gen3C~\cite{ren2025gen3c}     & \underline{21.93} & \underline{0.238} & \underline{0.925} & \textbf{0.949} & \underline{19.22} & \underline{0.303} & \underline{0.912} & \underline{0.906} \\
    & Voyager~\cite{huang2025voyager}   & 17.43 & 0.315 & 0.900 & 0.877 & 15.88 & 0.389 & 0.881 & 0.827 \\
    & VMem~\cite{li2025vmem}      & 11.69 & 0.626 & 0.812 & 0.547 & 11.07 & 0.678 & 0.754 & 0.402 \\
    & SPMem~\cite{wu2025video}     & 18.42 & 0.279 & 0.920 & 0.902 & 16.43 & 0.375 & 0.892 & 0.819 \\
    & \textbf{\method (ours)} & \textbf{22.46} & \textbf{0.168} & \textbf{0.965} & \textbf{0.949} & \textbf{19.34} & \textbf{0.254} & \textbf{0.952} & \textbf{0.910} \\
    \bottomrule
  \end{tabular}
  }
\end{table}

\subsection{Long-Context Generation under Large Viewpoint Change}
\label{sec:exp_long}

\begin{figure}[t]
  \centering
  \includegraphics[width=\linewidth]{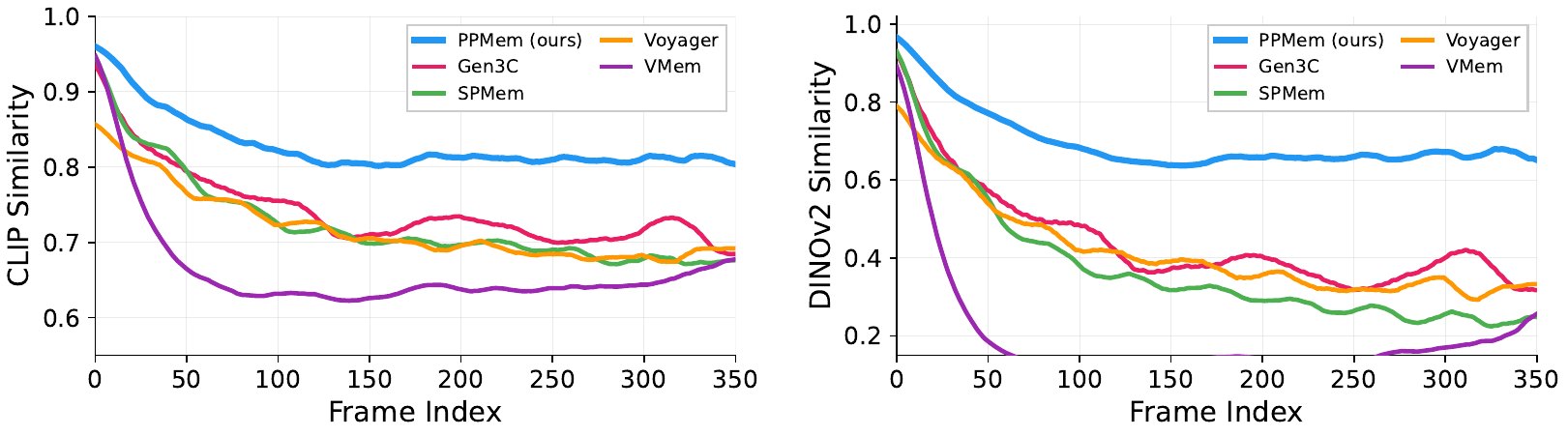}
  \caption{
    \textbf{Perceptual similarity vs.\ frame index on DL3DV long-context generation (${\sim}$350 frames).}
    Left: CLIP similarity. Right: DINOv2 similarity. \method (blue) maintains the highest perceptual fidelity across the full generation. VMem collapses early; Gen3C and SPMem degrade steadily; the gap widens with frame count, especially on DINOv2.
  }
  \label{fig:degradation}
\end{figure}

\begin{figure*}[t]
  \centering
  \includegraphics[width=\linewidth]{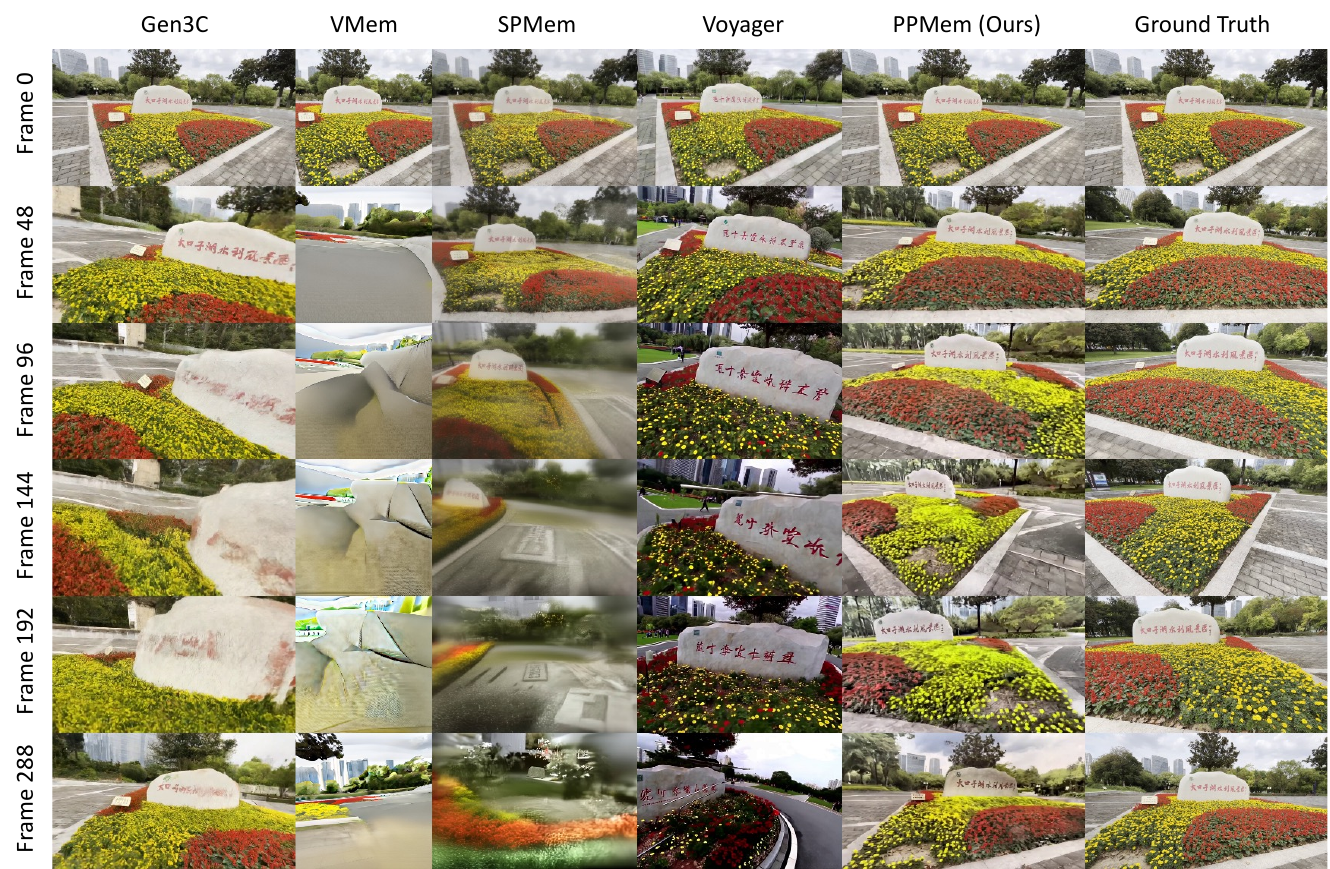}
  \caption{
    \textbf{Qualitative long-context comparison.}
    \method preserves scene structure and color consistency across the full sequence. VMem collapses early under large viewpoint change; SPMem grows vague and drifts; Gen3C and Voyager lose structural alignment as the viewpoint diverges.  }
  \label{fig:qualitative}
\end{figure*}

We evaluate all methods on the full DL3DV test sequences (300--400 frames each), where large camera motion across many autoregressive chunks tests each method's ability to maintain coherent generation far from the initial viewpoint. Figure~\ref{fig:degradation} plots CLIP and DINOv2 similarity against frame index: \method maintains the highest perceptual fidelity throughout, with the gap widening over time --- most dramatically on DINOv2, where \method leads the next-best method by ${\sim}$0.27 at frame 300. Figure~\ref{fig:qualitative} shows a representative example: \method preserves scene structure and color consistency across the full sequence, while all baselines degrade visibly as the viewpoint diverges.

\subsection{Ablations}
\label{sec:exp_ablations}

We ablate four design aspects below. All variants are trained on the DL3DV indoor subset at $288{\times}512$, evaluated on the 56-scene indoor test split over 3 autoregressive chunks, and share the same Stage-1 backbone and Stage-2 training recipe. Additional ablation studies on context frame composition and plane pool merging stages are in Appendix~\ref{app:ablations}.

\begin{table}[t]
  \centering
  \caption{\textbf{3D representation and prediction ablation.} We compare our joint plane pool against three inference-only alternatives (same checkpoint, no retraining): a same-model rollout without persistent memory, Gen3C-style per-frame MoGe-2 point cloud memory, and post-hoc standalone Plana3R reconstruction.}
  \label{tab:abl_repr}
  \small
  \resizebox{\linewidth}{!}{%
  \begin{tabular}{lccccccc}
    \toprule
    Variant & PSNR$\uparrow$ & LPIPS$\downarrow$ & CLIP$_\text{sim}\uparrow$ & DINOv2$_\text{sim}\uparrow$ & AbsRel$\downarrow$ & $\delta{<}1.25\uparrow$ & Normal$^{\circ}\downarrow$ \\
    \midrule
    Without persistent memory & 14.58 & 0.407 & \textbf{0.835} & 0.733 & 0.272 & 64.0 & 25.2 \\
    MoGe-2 point cloud    & 14.37 & 0.423 & 0.832 & 0.716 & 0.339 & 55.0 & 25.8 \\
    Post-hoc Plana3R       & 14.22 & 0.432 & 0.829 & 0.713 & 0.311 & 57.0 & 26.3 \\
    \textbf{Joint plane pool (ours)} & \textbf{14.91} & \textbf{0.387} & \textbf{0.835} & \textbf{0.741} & \textbf{0.210} & \textbf{70.3} & \textbf{24.1} \\
    \bottomrule
  \end{tabular}}
\end{table}

\paragraph{3D representation and prediction (Table~\ref{tab:abl_repr}).}
To validate the choice of plane primitives and joint prediction, we compare three alternative configurations at inference time (no retraining):
(i) \textbf{Without persistent memory}: a same-model rollout using the model's own predicted depth but no persistent memory.
(ii) \textbf{Point cloud memory}: we replace the plane pool with a Gen3C-style mechanism --- MoGe-2~\cite{wang2025moge2} estimates per-frame monocular depth on each generated frame, and the resulting point cloud conditions the next chunk via forward warping.
(iii) \textbf{Post-hoc Plana3R}: instead of using the joint plane prediction head, we run a standalone pretrained Plana3R model~\cite{liu2025plana3r} on each generated frame post-hoc and accumulate the resulting planes into the pool.
Our joint plane pool outperforms all three alternatives, with particularly large gains on depth ($\delta{<}1.25$: $70.3$ vs $64.0$/$55.0$/$57.0$), indicating that the persistent memory contributes primarily to geometric consistency and long-horizon stability.

\begin{figure}[t]
  \centering
  \begin{subfigure}[t]{0.63\linewidth}
    \centering
    \includegraphics[width=\linewidth]{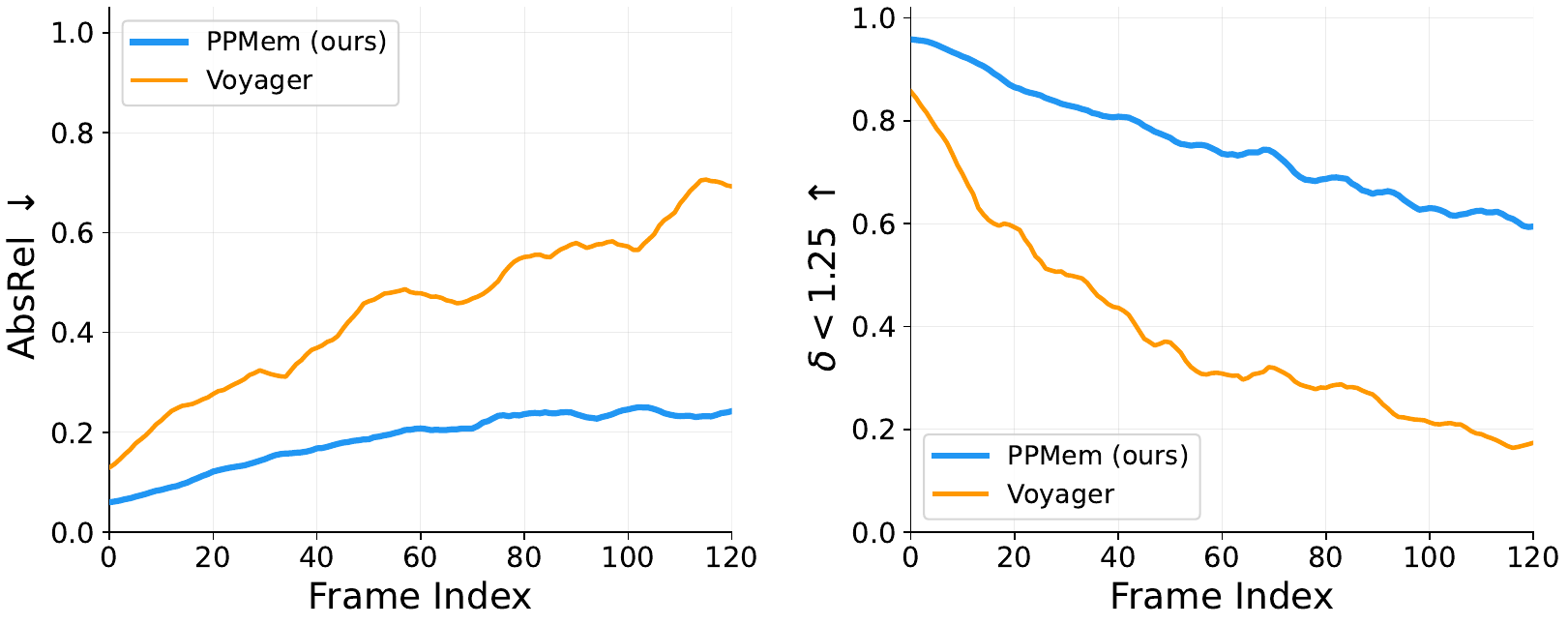}
    \caption{Per-frame depth accuracy on DL3DV (121 frames).}
    \label{fig:depth_curve}
  \end{subfigure}
  \hfill
  \begin{subfigure}[t]{0.35\linewidth}
    \centering
    \includegraphics[width=\linewidth]{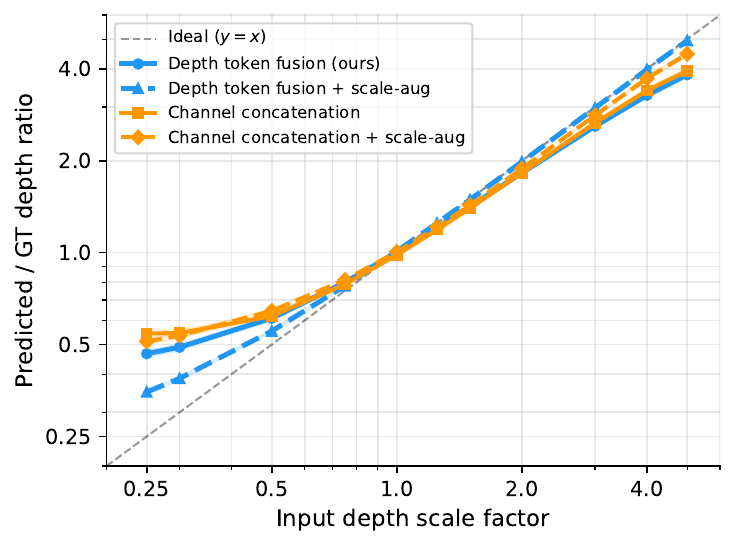}
    \caption{DTF scale faithfulness (log-log).}
    \label{fig:dtf_scale}
  \end{subfigure}
  \caption{
    \textbf{Depth quality analysis.}
    \emph{(a)} Per-frame depth metrics (AbsRel $\downarrow$, $\delta{<}1.25$ $\uparrow$) on DL3DV over 121 frames. \method maintains consistently lower depth error than Voyager, the only other baseline with joint depth output.
    \emph{(b)} We scale the input depth conditioning by a known factor and measure the predicted/GT depth ratio. Concat + DTF (ours) tracks input scale faithfully ($1.00 \pm 0.01$ at scale 1.0, mean $\pm$ s.e.\ over 56 scenes), while channel concatenation alone exhibits compressed response at extreme scales. Measured as deviation from the identity mapping over the full scale range, DTF also stays closer to it (natural-log RMSE $0.273$ vs.\ $0.320$ for concatenation alone; $0.131$ vs.\ $0.293$ with scale augmentation).
  }
  \label{fig:depth_and_dtf}
\end{figure}

\begin{table}[t]
  \centering
  \caption{\textbf{Depth conditioning mechanism ablation.} All variants share the same backbone and training data, varying only the depth conditioning mechanism.}
  \label{tab:abl_depth}
  \small
  \resizebox{\linewidth}{!}{%
  \begin{tabular}{lccccccc}
    \toprule
    Mechanism & PSNR$\uparrow$ & LPIPS$\downarrow$ & CLIP$_\text{sim}\uparrow$ & DINOv2$_\text{sim}\uparrow$ & AbsRel$\downarrow$ & $\delta{<}1.25\uparrow$ & Normal$^{\circ}\downarrow$ \\
    \midrule
    No depth conditioning           & 14.19 & 0.430 & 0.826 & 0.709 & 0.436 & 30.7 & 25.8 \\
    Channel concat only              & 14.88 & 0.388 & \textbf{0.838} & \textbf{0.743} & 0.212 & \textbf{72.3} & \textbf{23.8} \\
    \textbf{Concat + DTF (ours)}     & \textbf{14.91} & \textbf{0.387} & 0.835 & 0.741 & \textbf{0.210} & 70.3 & 24.1 \\
    \bottomrule
  \end{tabular}}
\end{table}

\paragraph{Depth conditioning mechanism (Table~\ref{tab:abl_depth}).}
The dominant effect is the presence of depth conditioning itself: without it, depth accuracy collapses (AbsRel $0.436$, $\delta{<}1.25 = 30.7\%$) and video quality also degrades (PSNR $14.19$ vs $14.91$). The video improvement is indirect: better depth prediction yields a higher-quality plane pool, which in turn provides better projected conditioning (RGB, masks, rendered depth) for subsequent chunks --- a cascading benefit across autoregressive steps. Given depth, channel concatenation and DTF perform comparably on the ablation benchmark, where training and evaluation both use well-aligned metric depth. At inference, the depth conditioning comes from plane pool rendering, which introduces holes, partial coverage, and scale imperfections. Figure~\ref{fig:dtf_scale} tests this via a scale faithfulness probe (whether predicted depth scales proportionally when the input depth conditioning is rescaled) --- we deliberately multiply the input depth by a known factor and measure the predicted/GT depth ratio. DTF tracks input scale faithfully ($1.00 \pm 0.01$ at scale 1.0), while channel concatenation exhibits a compressed response at extreme scales, suggesting that the VAE bottleneck discards fine-grained scale information. DTF's token-level fusion with explicit validity masking is better suited to the imperfect conditioning encountered during autoregressive inference. The two base variants are trained exclusively on well-aligned metric depth (scale ${\approx}\,1$). We additionally train both fusion schemes with per-video depth scale augmentation ($s \sim \mathrm{logU}[0.25, 4]$), and find that it substantially improves scale faithfulness in both cases, with DTF + scale augmentation tracking the input scale most closely overall (natural-log RMSE to the identity mapping: $0.131$, vs.\ $0.273$ for DTF alone and $0.293$ for concatenation with the same augmentation).

\begin{table}[!htb]
  \centering
  \caption{\textbf{Feature decoder and loss component ablations.} Top: decoder architecture (all else fixed). Bottom: loss removal (full model as baseline).}
  \label{tab:abl_decoder}
  \small
  \resizebox{\linewidth}{!}{%
  \begin{tabular}{lccccccc}
    \toprule
    Variant & PSNR$\uparrow$ & LPIPS$\downarrow$ & CLIP$_\text{sim}\uparrow$ & DINOv2$_\text{sim}\uparrow$ & AbsRel$\downarrow$ & $\delta{<}1.25\uparrow$ & Normal$^{\circ}\downarrow$ \\
    \midrule
    Plana3R upsampling head        & 14.68 & 0.395 & 0.832 & 0.737 & 0.256 & 59.2 & 26.2 \\
    \textbf{VAE-init decoder (ours)} & \textbf{14.91} & \textbf{0.387} & \textbf{0.835} & \textbf{0.741} & \textbf{0.210} & \textbf{70.3} & \textbf{24.1} \\
    \midrule
    $-$ rasterization loss & 14.82 & 0.391 & 0.834 & 0.736 & 0.217 & 66.4 & 25.4 \\
    $-$ normal loss & 13.27 & 0.502 & 0.807 & 0.675 & --- & --- & --- \\
    \bottomrule
  \end{tabular}}
\end{table}

\paragraph{Feature decoder and loss components (Table~\ref{tab:abl_decoder}).}
Replacing the original Plana3R upsampling head~\cite{liu2025plana3r} with a VAE-initialized decoder improves $\delta{<}1.25$ by $+11.1$ pp ($59.2 \to 70.3$) and AbsRel by $-0.046$ ($0.256 \to 0.210$), showing that the video VAE's learned spatial prior meaningfully benefits geometric prediction. For loss components, removing the rasterization loss degrades $\delta{<}1.25$ by $-3.9$ pp and Normal$^\circ$ by $+1.2^\circ$; removing the normal loss causes depth to collapse entirely (all-zero output), confirming it is critical for learning the coupled depth-normal parameterization.

\section{Conclusion}
\label{sec:conclusion}

We presented \method, a video world model that uses plane primitives as persistent 3D memory for long-context generation. The plane pool satisfies the four properties we identified for cross-chunk memory: it is compact (${\sim}$100K primitives for a full scene), geometrically explicit (metric depth, normals, exportable surfaces), cross-view consistent (jointly predicted by the generation backbone), and adaptive (refined via geometric merging as generation proceeds). Experiments on DL3DV and RealEstate10K show that \method sustains both video quality and depth accuracy under large viewpoint changes, outperforming Gen3C, Voyager, VMem, and SPMem.

\paragraph{Limitations and future work.}
\method is best suited for structured scenes where planar primitives are a good geometric prior. A quantitative breakdown of performance versus scene planarity is provided in Appendix~\ref{app:planarity}. For highly organic environments (dense vegetation, water) the representation may capture less detail than point-based alternatives. Extreme camera trajectories not seen during training can also degrade quality. Our two-stage design deliberately freezes the pretrained video backbone and trains geometry through lightweight adapters, which preserves the backbone's generation quality but forgoes end-to-end co-adaptation of the video and geometry objectives; jointly optimizing both is a more general alternative we leave to future work. Extending to dynamic plane primitives and scaling to open-world scenes are natural next steps. Like all video generation models, \method could potentially be misused for synthetic media creation.

\begin{ack}
This work was supported by the National Natural Science Foundation of China (62461160309) and NSFC-RGC Joint Research Scheme N\_HKU705/24.
\end{ack}

\bibliographystyle{plainnat}
\bibliography{ref}

\clearpage
\appendix

\renewcommand{\thetable}{S\arabic{table}}
\renewcommand{\thefigure}{S\arabic{figure}}
\setcounter{table}{0}
\setcounter{figure}{0}

\section{Implementation Details}
\label{app:impl}

\subsection{Dataset Preprocessing}
\label{app:data}

\paragraph{Training data.}
For ablation experiments (Tables~\ref{tab:abl_repr}--\ref{tab:abl_decoder} and Tables~\ref{tab:context}--\ref{tab:merge_phases}), we use 5K preprocessed DL3DV~\cite{ling2024dl3dv} indoor video clips at $\mathbf{288{\times}512}$ resolution. For the final model, we train on a mixture of DL3DV and RealEstate10K~\cite{zhou2018stereo} (80K+ clips) at $\mathbf{576{\times}1024}$ resolution. Each clip is paired with per-frame metric depth, surface normals, and camera intrinsics/extrinsics. Dense monocular depth and normals are estimated by MoGe-2~\cite{wang2025moge2} and aligned to multi-view reconstruction from VGGT~\cite{wang2025vggt} via per-frame scale fitting; the median scale is applied globally to place all depths and camera translations in a consistent metric coordinate system.

\paragraph{Offline conditioning pipeline.}
A central design principle of \method is that \emph{all training conditioning is generated by the same plane-based rendering process used at inference}, so that the model trains on realistic imperfections (holes, depth discontinuities, partial coverage) rather than perfect ground-truth projections. Concretely, we run an offline incremental plane reconstruction pipeline~\cite{tan2025planarsplatting,lipmap} on each video clip to produce all conditioning signals:

\begin{enumerate}
  \itemsep0em
  \item \textbf{Incremental reconstruction.} Frames are partitioned into segments of 24. Each segment undergoes plane-based reconstruction (optimizing plane primitives to fit the aligned metric depth). After each segment, new planes are merged with the accumulated pool and jointly optimized across all frames observed so far.
  \item \textbf{Conditioning generation.} After processing segment~$k$, we render the accumulated plane pool at the \emph{next} segment's cameras (segment~$k{+}1$), producing partial depth maps that serve as the depth conditioning for that segment's training sample. This one-step look-ahead mirrors the inference-time loop, where each chunk's depth conditioning comes from rendering the pool accumulated from all \emph{previous} chunks. Two source views are also selected from all preceding segments' frames via greedy union-coverage maximization and forward-warped to each target camera, producing projected RGB images and soft visibility masks (continuous values in $[0,1]$ from bilinear splatting weights).
  \item \textbf{Context frame selection.} Overlap frames are selected from all preceding segments via greedy union-coverage maximization over the current segment's target cameras. These are \emph{actual video frames} (not rendered views); their accompanying depth comes from the accumulated plane pool render at those cameras.
\end{enumerate}

Because the conditioning is rendered from the same plane-based reconstruction used at inference, it naturally exhibits the same artifact distribution: holes in unseen regions, depth edges at object boundaries, partial coverage from oblique viewpoints, and planar approximation errors. The gap between training and autoregressive inference thus reduces to the difference in plane quality (offline reconstruction vs.\ online model predictions), rather than a distribution shift between perfect ground truth and imperfect model outputs.

\paragraph{Context frames.}
For each training segment, three types of context frames are prepended along the temporal dimension:
\begin{itemize}
  \itemsep0em
  \item \textbf{Anchor}: frame~0 of the video, providing a long-range identity anchor.
  \item \textbf{History}: real RGB frames stratified-randomly sampled from preceding segments, each accompanied by depth rendered from the accumulated plane pool.
  \item \textbf{Overlap}: real RGB frames selected by greedy union-coverage maximization (as described above), with plane-pool-rendered depth.
\end{itemize}
Each context type is assigned a distinct negative RoPE temporal index, encoding it as temporally prior to the generated sequence. Context dropout is applied with probability 0.1 during training (the entire context block is dropped), encouraging robustness to partial or absent context at inference. We ablate the contribution of each context type in Table~\ref{tab:context}.

\paragraph{DiT input assembly.}
The DiT receives conditioning through three complementary pathways:

\emph{Channel concatenation.} All RGB, mask, and depth signals are VAE-encoded using the Wan~2.2 separate VAE (1:1 frame mapping, $z_\text{dim}{=}48$). Six such latents are concatenated along the channel dimension: the noisy video latent, two projected source-view RGB latents, two projected visibility-mask latents, and the depth conditioning latent, yielding $6 {\times} 48 = 288$ input channels in total.

\emph{Temporal prepend.} Context frames (anchor, history, overlap) are prepended as clean tokens (timestep ${=}\,0$) before the main frames along the temporal dimension, providing the model with real observed frames as conditioning memory.

\emph{Token-level fusion.} The same depth map also enters the model at pixel resolution through the DTF module (\S\ref{sec:method_dtf}), providing fine-grained geometric awareness via self-attention between depth tokens and DiT feature tokens.

\subsection{Training Hyperparameters}
\label{app:train_hp}

\begin{table}[h]
  \centering
  \caption{\textbf{Full training hyperparameters for Stage 2.} Stage 2 freezes the DiT backbone and injects LoRA adapters on self-attention; only geometry-specific modules and LoRA are trained. Stage 1 hyperparameters are similar except no plane loss, no DTF, no freeze/LoRA.}
  \label{tab:hp_full}
  \small
  \begin{tabular}{lll}
    \toprule
    Category & Hyperparameter & Value \\
    \midrule
    \multirow{7}{*}{Model}
      & Base video DiT                  & Wan~2.2 TI2V 5B \\
      & VAE                             & Per-frame (1:1 frame mapping, $z_\text{dim}{=}48$) \\
      & Depth token fusion encoder      & ViT-Base (${\sim}$87M params), DINOv2-Base init \\
      & Feature decoder                 & Initialized from Wan VAE upsample block \\
      & LoRA (Stage 2)                  & Rank 256, $\alpha{=}256$, self-attention, all 30 blocks \\
      & Trainable params (Stage 2)      & ${\sim}$377M / 5.4B (7.0\%) \\
      & Precision                       & bfloat16 \\
    \midrule
    \multirow{5}{*}{Optimization}
      & Optimizer             & AdamW \\
      & Learning rate         & $1\times10^{-5}$ \\
      & LR warmup             & 50 iterations \\
      & Total iterations (Stage 2) & 5{,}000 \\
      & Gradient clip norm    & 2.0 \\
    \midrule
    \multirow{4}{*}{Data}
      & Frames per chunk             & 25 \\
      & Resolution                   & $576 \times 1024$ (final) / $288 \times 512$ (ablation) \\
      & Classifier-free guidance dropout  & 0.10 \\
      & Context frame dropout             & 0.10 \\
    \midrule
    \multirow{6}{*}{Loss}
      & $\lambda_d$ (per-patch depth)          & 0.2 \\
      & $\lambda_n$ (per-patch normal)         & 0.05 \\
      & $\lambda_d^r$ (rasterization depth)    & 0.02 \\
      & $\lambda_n^r$ (rasterization normal)   & 0.01 \\
      & $\lambda_s$ (rendered depth smoothness) & 0.02 \\
      & Rasterization loss warmup              & 1000 iterations \\
    \midrule
    \multirow{2}{*}{\makecell[l]{Depth token\\fusion}}
      & Mixed-patch drop probability    & 0.75 \\
      & Target overall mask ratio       & $[0.6,\, 0.9]$ \\
    \bottomrule
  \end{tabular}
\end{table}

\paragraph{Plane head initialization.}
Each of the three prediction MLPs (center depth, rectangular extents, rotation quaternion) is initialized so that the last layer produces a stable constant output before training begins: depth bias is set to 2.0\,m with zero weights, radii bias to 0.1\,m with zero weights, and rotation bias to the identity quaternion $[1,0,0,0]$ with near-zero weights ($\sigma{=}0.001$). During the rasterization-loss warmup (first 1000 iterations), depth and normals are supervised by per-patch losses; the extents receive no direct gradient but remain at their initialized constant, ensuring stable rasterization once the rendering loss is enabled.

\subsection{Inference Setup}
\label{app:inference}

\method generates video autoregressively in 25-frame chunks; the last frame of each chunk is reused as the first frame of the next. Given an input image, a text prompt, and a target camera trajectory, the full pipeline proceeds as follows.

\begin{algorithm}[h]
  \caption{\method full inference loop.}
  \label{alg:inference}
  \begin{algorithmic}[1]
    \Require input image $I_0$, text prompt, target camera trajectory, supplied as camera-control inputs in fixed-length chunks (the last frame of each chunk seeds the next).
    \State \textbf{Initialize}: plane pool $\mathcal{P} \leftarrow \emptyset$. Estimate depth for $I_0$ (monocular estimator, or ground-truth depth if available) and project it to chunk 0's target cameras as the initial depth conditioning ($I_0$ serves as both source views).
    \For{chunk $k = 0, 1, 2, \ldots$}
      \State \textbf{Assemble conditioning}: two coverage-selected source views projected to the target cameras (RGB + masks), context frames (anchor, history, overlap), and the chunk's partial depth maps --- VAE-encoded for latent channel concatenation and raw for DTF.
      \State \textbf{Video denoising}: run the denoising schedule with the frozen Stage-1 backbone under classifier-free guidance $\rightarrow$ video chunk $V_k$.
      \State \textbf{Geometry pass}: re-run the final denoising step with the Stage-2 geometry LoRA enabled; the plane prediction head regresses plane primitives from the LoRA-adapted features; transform them to world space and rasterize them into per-frame depth/normal maps.
      \State \textbf{Pool update and next-chunk render}: merge the predicted planes into $\mathcal{P}$ (Algorithm~\ref{alg:merge}); render the updated $\mathcal{P}$ at chunk $k{+}1$'s target cameras via rectangular Gaussian rasterization $\rightarrow$ partial depth maps for the next iteration.
    \EndFor
    \State \textbf{Output}: generated video, per-frame depth/normal maps, final pool $\mathcal{P}$.
  \end{algorithmic}
\end{algorithm}

\paragraph{Initial chunk (chunk 0).}
The input image is used as the first frame. An initial depth estimate (from any monocular depth estimator or ground-truth depth, if available) is projected to the remaining 24 target cameras to produce partial RGB images and visibility masks (the single input frame is used as both source views to match the two-view conditioning format). The initial depth also serves as the depth conditioning signal for DTF. The DiT then generates the first 25-frame video clip following the same two-pass generation procedure described below (50 denoising steps with the frozen backbone, then one geometry LoRA pass); the predicted plane primitives initialize the plane pool.

\paragraph{Subsequent chunks (chunk $k \geq 1$).}
Each subsequent chunk follows a three-phase pipeline:
\begin{enumerate}
  \itemsep0em
  \item \textbf{Conditioning}: the cumulative plane pool is rendered at the next chunk's target cameras via rectangular Gaussian rasterization, producing partial depth maps. Two source views are selected from the pool's accumulated frames (greedy coverage maximization) and projected to the target cameras. Context frames --- an anchor frame (frame 0 of the video), up to 5 history frames from preceding chunks, and coverage-sorted overlap frames --- are prepended along the temporal dimension.
  \item \textbf{Generation and plane prediction}: the DiT performs 50 denoising steps with classifier-free guidance scale 5.0, producing the generated video. We then re-run the final denoising step with the Stage~2 LoRA adapters enabled: this additional forward pass provides geometry-adapted DiT features to the plane prediction head, which regresses plane primitives. The 50-step video generation uses only the frozen Stage~1 backbone, preserving video quality; the extra LoRA step serves solely to produce plane predictions. Each predicted plane is transformed from camera space to world space using the known camera extrinsics.
  \item \textbf{Pool update}: the predicted planes are merged into the cumulative pool via the merging pipeline (Algorithm~\ref{alg:merge}), which removes floaters, deduplicates redundant planes, and prunes planes that will be better observed by the next chunk.
\end{enumerate}

\paragraph{Output.}
After all chunks are generated, \method produces: (a) the full generated video, (b) per-frame metric-scale depth and surface normal maps (from the plane prediction head), and (c) the cumulative plane pool as a compact 3D reconstruction of the generated scene (exportable as a mesh via triangulation of the rectangular primitives).

\paragraph{Compute overhead.}
The plane pool adds negligible cost relative to video denoising. At $576{\times}1024$, denoising dominates each chunk (${\sim}$128\,s); pool rasterization (${\sim}$1.7\,s for 25 views) and merging (${\sim}$0.3\,s) are two orders of magnitude cheaper.

\paragraph{End-to-end memory accounting.}
We account separately for scene-growing, bounded per-chunk, and fixed model
state. At a long-rollout snapshot, our persistent pool (106K planes, 4.8\,MiB
including merger metadata) is substantially smaller than a matched point-cloud
variant that accumulates per-pixel 3D points over the same rollout and
resolution (1.7\,GiB). Context frames, projected views, rendered depth, and
merge buffers remain bounded per chunk, with a maximum simultaneous non-model
state of 41\,MiB, including the plane pool. DTF, the geometry decoder/head, and
geometry LoRA add 1.1\,GiB of fixed BF16 model state; no separate acceleration
structure is used.

\subsection{Plane Pool Merging Algorithm}
\label{app:merge}

Algorithm~\ref{alg:merge} details the full merging pipeline. Each plane carries an \emph{observation quality score} $s = r / |\cos\theta|$ (defined in main paper \S\ref{sec:method_repr}), where $r$ is the Euclidean distance to the creating camera and $\theta$ is the angle between the viewing ray and the surface normal. A lower score means the plane was observed from a closer, more frontal viewpoint and is therefore more reliable. This score drives the quality comparisons in stages 3, 5, and 6.

\begin{algorithm}[h]
  \caption{Plane pool merging pipeline.}
  \label{alg:merge}
  \begin{algorithmic}[1]
    \Require Pool $\mathcal{P}$, new planes $\mathcal{P}_{\text{new}}$, chunk cameras $\{(\mathbf{K}_t, \mathbf{E}_t)\}$, rendered depth $\mathbf{D}_t$, next-chunk cameras.
    \Statex
    \State \textbf{1.\ Floater removal} (on $\mathcal{P}$): delete pool planes whose depth $<$ new depth $\mathbf{D}_t$ by margin $\delta_{\text{occ}}$, if ${\geq}\rho$ cameras agree.
    \State \textbf{2.\ Boundary rejection} (on $\mathcal{P}_{\text{new}}$): reject planes at depth discontinuities (Sobel gradient on patch grid $> \theta_e$).
    \State \textbf{3.\ Within-chunk dedup} (on $\mathcal{P}_{\text{new}}$): among nearby planes ($<\epsilon_{\text{same}}$) within the chunk, keep only the one with best quality score $s$. Removes ${\sim}80\%$ of per-chunk planes.
    \State \textbf{4.\ Distance filter} (on $\mathcal{P}_{\text{new}}$): reject planes farther than $d_{\max}$ from creating camera, unless pool has no coverage there.
    \State \textbf{5.\ Cross-chunk replacement} (on $\mathcal{P} \cup \mathcal{P}_{\text{new}}$): among nearby pairs across pool and new chunk, keep the better-scored plane.
    \State Append survivors to $\mathcal{P}$ with metadata.
    \State \textbf{6.\ Forward-looking prune} (on $\mathcal{P}$): remove planes that will be viewed much closer in the next chunk ($r_{\text{next}} / r_{\text{created}} < \tau$).
    \State \Return $\mathcal{P}$.
  \end{algorithmic}
\end{algorithm}

Default thresholds: $\delta_{\text{occ}}{=}0.1$ (10\% relative margin), $\theta_e{=}1.0$ (Sobel on log-depth), $\epsilon_{\text{same}}{=}0.5$\,m, $d_{\max}{=}5$\,m, $\rho{=}0.2$ (20\% camera agreement), $\tau{=}0.3$.

We ablate the contribution of each stage in Table~\ref{tab:merge_phases} (Section~\ref{app:merge_phases}).

\section{Additional Ablations and Comparisons}
\label{app:ablations}

\subsection{Context Configurations}
\label{app:context}

Table~\ref{tab:context} ablates the three types of context frames that condition each autoregressive chunk: (i) \emph{overlap}: past RGB frames selected by coverage of the target viewpoints; (ii) \emph{history} frames from preceding chunks; and (iii) an \emph{anchor} frame (frame 0 of the video). All configurations are evaluated at inference time without retraining, leveraging the 10\% context dropout applied during training.

\paragraph{Component ablation (top).} Removing all context degrades $\delta{<}1.25$ from 70.3 to 64.2. Overlap frames alone recover most of this gap ($+$5.0\,pp) and achieve the lowest AbsRel (0.208). Adding history and anchor further improves PSNR and perceptual metrics, with the full combination achieving the best balance across all metrics.

\paragraph{History scaling (bottom).} Increasing the number of history frames beyond the default (5) yields only marginal gains (PSNR 14.91 $\to$ 14.94) while prefix sequence length and attention cost grow with the history count. We keep 5 as the default for this quality/compute tradeoff.
\begin{table}[h]
  \centering
  \caption{\textbf{Context frame configuration ablation, inference-only.} We vary the context frame composition at inference time without retraining (the model was trained with context dropout 0.1). Anchor = frame 0 of the video; History = frames from preceding chunks; Overlap = past RGB frames selected by coverage of the target viewpoints. Top: component ablation. Bottom: history count scaling. Evaluated on 56 DL3DV indoor scenes at $288{\times}512$.}
  \label{tab:context}
  \small
  \resizebox{\linewidth}{!}{%
  \begin{tabular}{lccc ccccccc}
    \toprule
    & Anchor & History & Overlap & PSNR$\uparrow$ & LPIPS$\downarrow$ & CLIP$_\text{sim}\uparrow$ & DINOv2$_\text{sim}\uparrow$ & AbsRel$\downarrow$ & $\delta{<}1.25\uparrow$ & Normal$^{\circ}\downarrow$ \\
    \midrule
    \multicolumn{11}{l}{\emph{Component ablation}} \\
    No context              & \ding{55} & \ding{55} & \ding{55} & 14.60 & 0.403 & \textbf{0.835} & 0.730 & 0.219 & 64.2 & 24.5 \\
    Overlap only            & \ding{55} & \ding{55} & \ding{51} & 14.73 & 0.394 & \underline{0.834} & 0.738 & \textbf{0.208} & 69.2 & 24.3 \\
    History + Overlap       & \ding{55} & \ding{51} & \ding{51} & 14.77 & 0.393 & 0.832 & \underline{0.739} & 0.218 & \underline{70.2} & 24.3 \\
    \textbf{All (default)}  & \ding{51} & \ding{51} & \ding{51} & \underline{14.91} & \textbf{0.387} & \textbf{0.835} & \textbf{0.741} & \underline{0.210} & \textbf{70.3} & \underline{24.1} \\
    \midrule
    \multicolumn{11}{l}{\emph{History scaling (Anchor + Overlap + varying history count)}} \\
    Hist = 5 (default)  & \ding{51} & \ding{51} & \ding{51} & \underline{14.91} & \textbf{0.387} & \textbf{0.835} & \textbf{0.741} & \underline{0.210} & \textbf{70.3} & \underline{24.1} \\
    Hist = 10           & \ding{51} & \ding{51} & \ding{51} & 14.80 & 0.393 & 0.833 & 0.735 & 0.220 & 69.0 & 24.2 \\
    Hist = 20           & \ding{51} & \ding{51} & \ding{51} & 14.82 & 0.393 & \underline{0.834} & 0.733 & 0.225 & 68.6 & 24.2 \\
    Hist = 36           & \ding{51} & \ding{51} & \ding{51} & \textbf{14.94} & \underline{0.390} & 0.833 & 0.732 & 0.227 & 67.9 & \textbf{24.0} \\
    \bottomrule
  \end{tabular}}
\end{table}

\subsection{Plane Pool Merging Phase Ablation}
\label{app:merge_phases}

Table~\ref{tab:merge_phases} cumulatively adds merging mechanisms to the baseline. Within-chunk dedup (stage~3) provides the largest single improvement, reducing per-chunk plane count by ${\sim}$4.5$\times$ (518K $\to$ 116K) while improving $\delta{<}1.25$ on both indoor and outdoor scenes. Every subsequent addition further improves $\delta{<}1.25$ monotonically on both splits. Boundary rejection uses a single global threshold shared across indoor and outdoor scenes, trading a small amount of coverage for stricter geometry filtering.

\begin{table}[h]
  \centering
  \caption{\textbf{Progressive build-up of the plane merging pipeline.} Depth accuracy is measured by absolute relative error (Abs Rel) and the percentage of pixels with predicted/GT depth ratio within 1.25 ($\delta{<}1.25$). Coverage is the fraction of pixels receiving a valid rendered depth. Evaluated on 16 DL3DV scenes (8 outdoor + 8 indoor).}
  \label{tab:merge_phases}
  \small
  \resizebox{\linewidth}{!}{%
  \begin{tabular}{lccccccc}
    \toprule
    & \multicolumn{3}{c}{Outdoor (8 scenes)} & \multicolumn{3}{c}{Indoor (8 scenes)} & \\
    \cmidrule(lr){2-4}\cmidrule(lr){5-7}
    Mechanism (cumulative) & Abs Rel$\downarrow$ & $\delta{<}1.25\uparrow$ & Coverage & Abs Rel$\downarrow$ & $\delta{<}1.25\uparrow$ & Coverage & Planes \\
    \midrule
    Baseline (no merging improvements) & 0.479 & 41.1 & 0.865 & 0.311 & 46.6 & 0.948 & 518K \\
    \;+ Within-chunk dedup (stage 3) & 0.486 & 44.7 & 0.833 & 0.297 & 51.8 & 0.920 & 116K \\
    \;+ Cross-chunk replacement (stage 5) & 0.477 & 47.0 & 0.803 & 0.296 & 52.7 & 0.916 & 100K \\
    \;+ Forward-looking prune (stage 6) & \textbf{0.469} & 48.3 & 0.798 & 0.285 & 53.9 & 0.908 & 100K \\
    \textbf{\;+ Boundary rejection (stage 2)} & 0.472 & \textbf{49.6} & 0.786 & \textbf{0.284} & \textbf{56.5} & 0.900 & \textbf{97K} \\
    \bottomrule
  \end{tabular}}
\end{table}

\begin{figure*}[t]
  \centering
  \includegraphics[width=\linewidth]{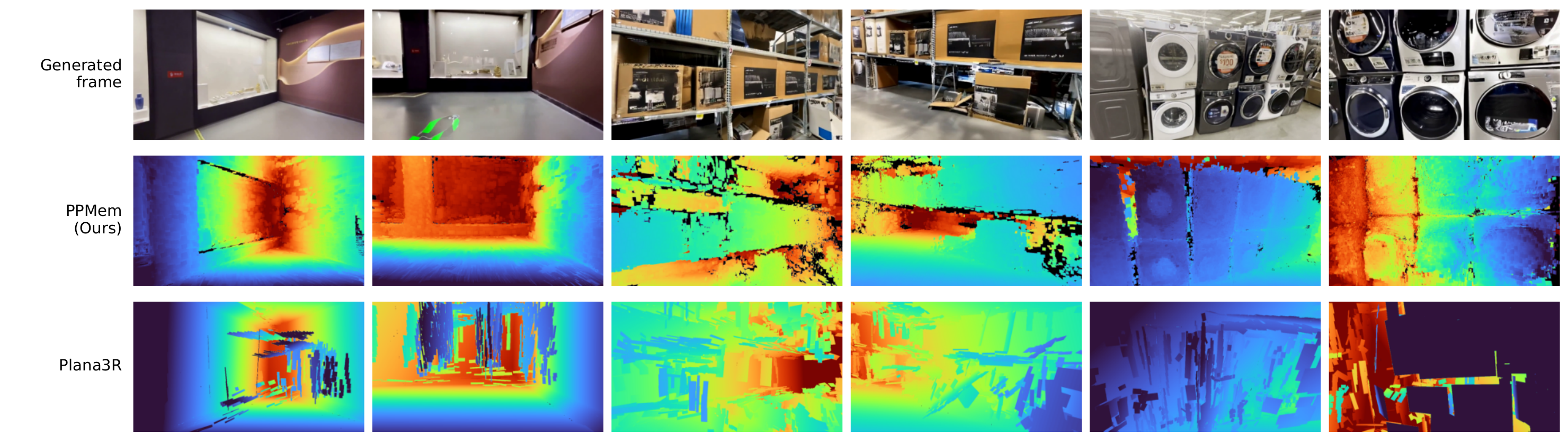}
  \caption{\textbf{Accumulated plane pool vs.\ Plana3R reconstruction.}
  We render depth from our plane pool and from planes reconstructed by
  Plana3R~\cite{liu2025plana3r} on the same generated video. The pool built online
  during generation stays consistent, whereas pairwise reconstructions conflict
  across views and fragment.}
  \label{fig:pool_vs_plana3r}
\end{figure*}

\paragraph{Accumulated plane pool.}
We render each final pool over the complete generated trajectory. Across 56 scenes, the accumulated pool compresses the per-chunk plane predictions into only ${\sim}$100K planes while retaining accuracy close to them (0.241 vs.\ 0.210 AbsRel), with 98.7\% image coverage.
Figure~\ref{fig:pool_vs_plana3r} compares the pool against Plana3R reconstructions of the same generated videos.

\subsection{Scene Planarity}
\label{app:planarity}

We compute scene planarity independently of \method predictions. For every view
in the reference sequence, we use estimated depth and surface normals to group
neighboring valid pixels satisfying surface-normal and local 3D plane-consistency
constraints. Scene planarity is the fraction of valid-depth pixels, pooled across
views, covered by connected planar regions of at least 256 pixels. Across the
100 DL3DV scenes, planar coverage has a Spearman correlation of 0.33 with PSNR
at 121 frames. Table~\ref{tab:planarity} groups the scenes into within-domain
tertiles of planar coverage.

\begin{table}[h]
  \centering
  \caption{\textbf{Performance by scene planarity.} The 100 DL3DV scenes are grouped into within-domain tertiles of planar coverage; all metrics are computed at $N{=}121$.}
  \label{tab:planarity}
  \small
  \begin{tabular}{lccc}
    \toprule
    Planar coverage & Low & Middle & High \\
    \midrule
    \# Scenes & 34 & 32 & 34 \\
    PSNR$\uparrow$        & 12.98 & 13.67 & 15.27 \\
    LPIPS$\downarrow$     & 0.482 & 0.466 & 0.443 \\
    AbsRel$\downarrow$    & 0.289 & 0.250 & 0.195 \\
    $\delta{<}1.25\uparrow$ & 64.2 & 72.5 & 79.0 \\
    \bottomrule
  \end{tabular}
\end{table}

Overall, lower planarity is associated with a gradual decrease in PSNR and
depth accuracy, while LPIPS varies only mildly across the three groups.

\subsection{Comparison to Non-3D-Persistent Baselines}
\label{app:non3d}

We additionally evaluate WorldPlay~\cite{sun2025worldplay} and
MemCam~\cite{gao2026memcam}, which use historical-frame context without an
explicit persistent 3D representation, on the same DL3DV evaluation set (Table~\ref{tab:non3d}).

\begin{table}[h]
  \centering
  \caption{\textbf{Comparison to non-3D-persistent baselines on DL3DV.} WorldPlay and MemCam condition on historical frames without an explicit persistent 3D representation. The evaluation set and frame budgets match Table~\ref{tab:short_range}.}
  \label{tab:non3d}
  \small
  \setlength{\tabcolsep}{4pt}
  \resizebox{\linewidth}{!}{%
  \begin{tabular}{l cccc cccc}
    \toprule
    & \multicolumn{4}{c}{$N{=}49$} & \multicolumn{4}{c}{$N{=}121$} \\
    \cmidrule(lr){2-5}\cmidrule(lr){6-9}
    Method & PSNR$\uparrow$ & LPIPS$\downarrow$ & CLIP$_\text{sim}\uparrow$ & DINOv2$_\text{sim}\uparrow$ & PSNR$\uparrow$ & LPIPS$\downarrow$ & CLIP$_\text{sim}\uparrow$ & DINOv2$_\text{sim}\uparrow$ \\
    \midrule
    WorldPlay~\cite{sun2025worldplay} & 13.22 & 0.532 & 0.883 & 0.784 & 11.98 & 0.604 & 0.848 & 0.683 \\
    MemCam~\cite{gao2026memcam}       & 12.35 & 0.592 & 0.782 & 0.614 & 11.08 & 0.657 & 0.732 & 0.472 \\
    \textbf{\method (ours)} & \textbf{16.99} & \textbf{0.335} & \textbf{0.909} & \textbf{0.861} & \textbf{13.98} & \textbf{0.464} & \textbf{0.863} & \textbf{0.769} \\
    \bottomrule
  \end{tabular}}
\end{table}

\section{Additional Qualitative Results}
\label{app:qualitative}

\begin{figure}[!htb]
  \centering
  \includegraphics[width=\linewidth]{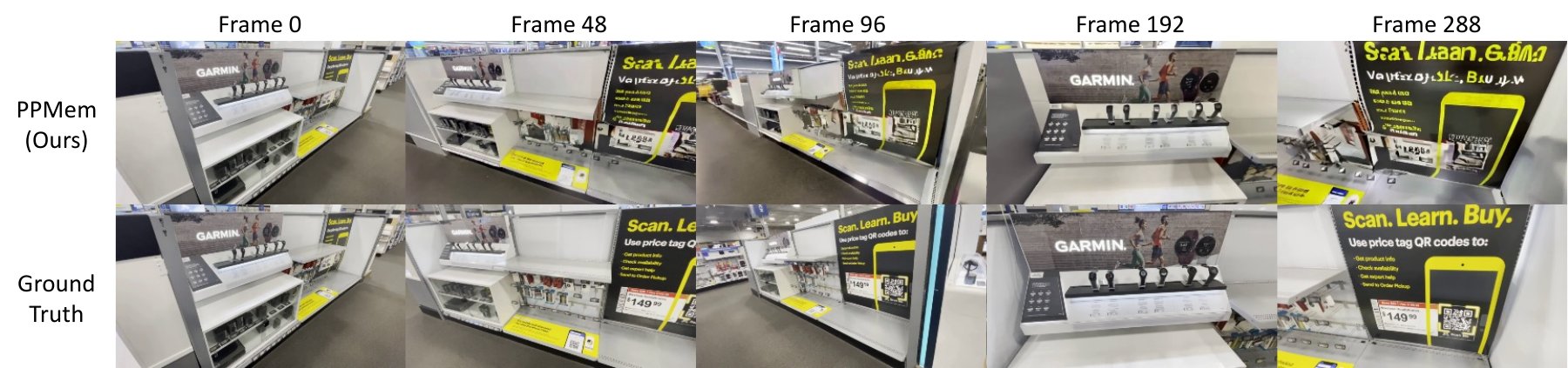}
  \caption{
    \textbf{Failure case: fine text rendering under zoom-in.}
    Scene structure and prominently visible text are preserved across 288 frames, but text that is barely legible in the initial viewpoint becomes garbled as the camera approaches. Faithful text synthesis from low-resolution observations remains an open challenge for video diffusion models.
  }
  \label{fig:failure}
\end{figure}

\begin{figure}[p]
  \centering
  \includegraphics[width=\linewidth]{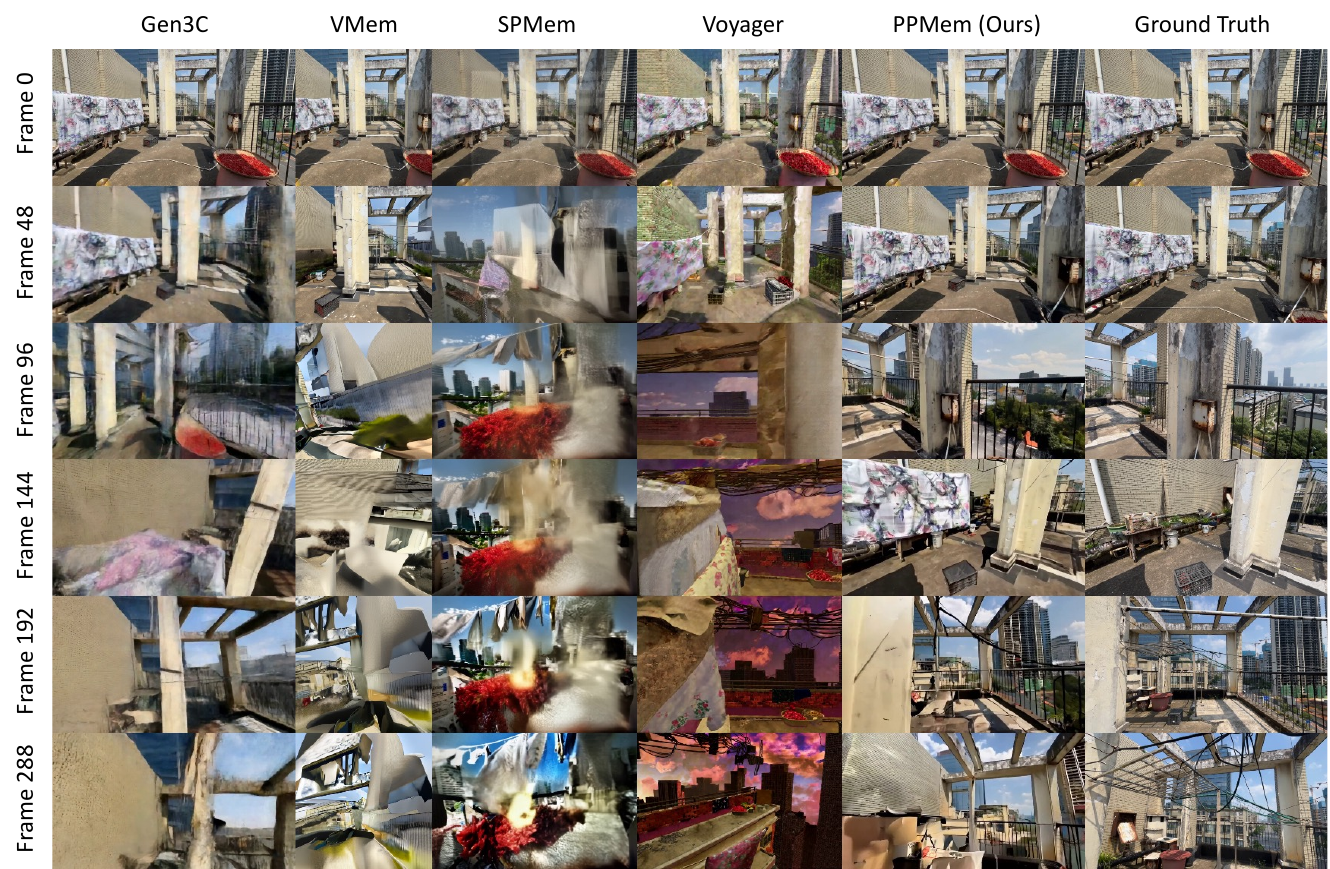}
  \caption{\textbf{Additional qualitative comparison: outdoor rooftop scene.} \method maintains structural consistency across 288 frames under large viewpoint change.}
  \label{fig:qualitative_rooftop}

  \vspace{24pt}
  \includegraphics[width=\linewidth]{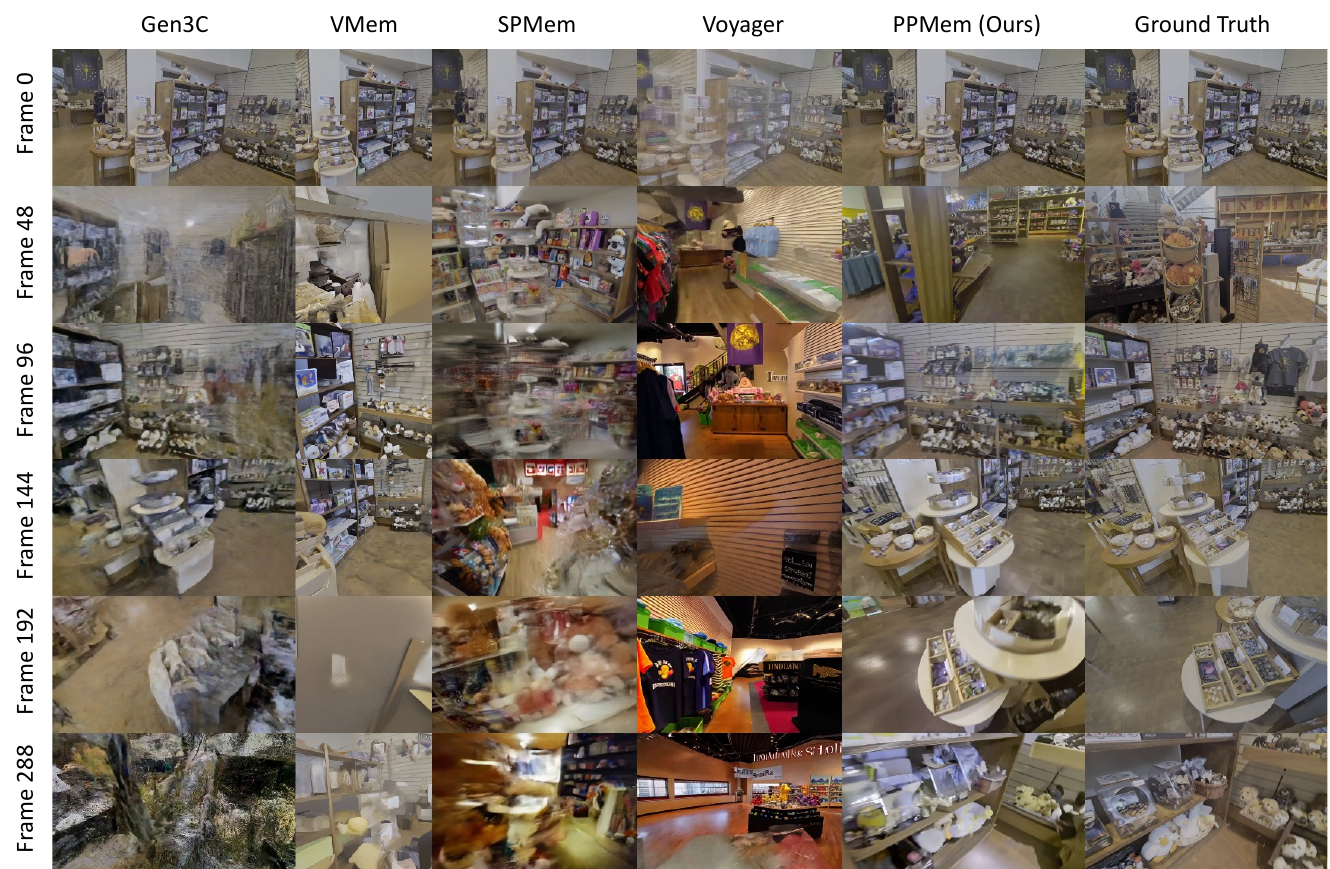}
  \caption{\textbf{Additional qualitative comparison: indoor store scene.} \method preserves scene layout and object identity while all baselines degrade.}
  \label{fig:qualitative_store}
\end{figure}

Figures~\ref{fig:qualitative_rooftop} and~\ref{fig:qualitative_store} compare long-context generation on DL3DV outdoor rooftop and indoor store scenes. Figure~\ref{fig:failure} shows degradation of fine text under zoom-in.

\clearpage
\makeatletter
\setlength{\@fptop}{0pt}
\setlength{\@fpbot}{0pt plus 1fil}
\makeatother

\end{document}